\documentclass{article}

\usepackage[preprint]{neurips_2022}

\usepackage[utf8]{inputenc} 
\usepackage[T1]{fontenc}    
\usepackage{hyperref}       
\usepackage{url}            
\usepackage{booktabs}       
\usepackage{amsfonts}       
\usepackage{nicefrac}       
\usepackage{microtype}      
\usepackage{soul}   
\usepackage[dvipsnames]{xcolor}
\usepackage{fancyvrb}
\usepackage{graphicx}
\usepackage{amsmath}
\usepackage{amssymb}
\usepackage{amsthm}
\usepackage{listings}
\usepackage{tikz}

\newcommand{\davinci}{\textit{Code-davinci-002}}

\title{ProofNet: Autoformalizing and Formally Proving Undergraduate-Level Mathematics}

\newcommand*\samethanks[1][\value{footnote}]{\footnotemark[#1]}

\iffalse
    \newcommand{\za}[1]{\textcolor{green}{}}
\else
    \newcommand{\za}[1]{\textcolor{violet}{\bf\small [ZA: #1]}}
\fi

\author{
  Zhangir Azerbayev\\
  Yale College\thanks{Work completed while at Carnegie Mellon University} \\
  \texttt{zhangir.azerbayev@yale.edu} \\
  \And 
  Bartosz Piotrowski \\
  University of Warsaw\samethanks \\
  \texttt{bartoszpiotrowski@post.pl}\\
  \And 
  Hailey Schoelkopf \\
  EleutherAI, Yale College \\
  \texttt{hailey.schoelkopf@yale.edu}\\
  \And
  Edward W. Ayers \\
  Carnegie Mellon University\\
  \texttt{contact@edayers.com}\\
  \And 
  Dragomir Radev \\
  Yale University \\
  \texttt{dragomir.radev@yale.edu}
  \And 
  Jeremy Avigad \\
  Carnegie Mellon University \\
  \texttt{avigad@cmu.edu}
}

\lstdefinestyle{mystyle}{
    inputencoding=utf8,
    basicstyle=\ttfamily\footnotesize,
    breakatwhitespace=false,         
    breaklines=true,                 
    captionpos=b,                    
    keepspaces=true,                 
    numbersep=5pt,                  
    showspaces=false,                
    showstringspaces=false,
    showtabs=false,                  
    tabsize=2, 
    frame=single
}

\begin{document}

\maketitle

\begin{abstract}
  We introduce \textsf{ProofNet}, a benchmark for autoformalization and formal proving of undergraduate-level mathematics. The \textsf{ProofNet} benchmarks consists of 371 examples, each consisting of a formal theorem statement in Lean 3, a natural language theorem statement, and a natural language proof. The problems are primarily drawn from popular undergraduate pure mathematics textbooks and cover topics such as real and complex analysis, linear algebra, abstract algebra, and topology. We intend for \textsf{ProofNet} to be a challenging benchmark that will drive progress in autoformalization and automatic theorem proving.  We report baseline results on statement autoformalization via in-context learning. Moreover, we introduce two novel statement autoformalization methods: \textit{prompt retrieval} and \textit{distilled backtranslation}. 
\end{abstract}

\section{Introduction}

\begin{figure*}[t]
\noindent
\begin{minipage}{0.50\textwidth}
\begin{center}
\large{Lean {\sf mathlib}}
\end{center}
\begin{flushleft}
    {\bf Formal theorem statement:}  \\
    \texttt{{\color{blue}theorem} {\color{orange}exists\_subgroup\_card\_pow\_prime}  \\
    \quad [fintype G] (p : $\mathbb{N}$) \{n : $\mathbb{N}$\} \\
    \quad [fact p.prime] (hdvd: p \^{} n | card G) : \\
    \quad $\exists$ K : subgroup G, fintype.card K = p\^{}n \\} 
\end{flushleft}
\end{minipage}\vline\phantom{v}\hfill%
\begin{minipage}{0.50\textwidth} 
\begin{flushleft}
\begin{center}
\large{{\sf ProofNet} dataset (ours)}
\end{center}
\vspace{5pt}

{\bf Formal theorem statement:} \\
\texttt{{\color{blue} theorem} {\color{orange}exercise\_4\_5\_14} \{G : {\color{blue}Type}*\} \\
\quad [group G] [fintype G] \\
\quad (hG : card G = {\color{ForestGreen}312}) : \\
\quad $\exists$ (p : $\mathbb{N}$) (P : sylow p G), P.normal} \\
\vspace{5pt}
{\bf Natural language theorem statement: } \\
Prove that a group of order 312 has a normal Sylow $p$-subgroup for some prime $p$ dividing its order. \\
\vspace{5pt}
\textbf{Natural language proof: } 
\end{flushleft}
    \begin{proof}
    Since $|G|=351=3^{2}.13$, $G$ has $3$-Sylow subgroup of order $9$, as well as $13$-Sylow subgroup of order $13$. Now, we count the number of such subgroups. Let $n_{13}$ be the number of $13$-Sylow subgroup and $n_{3}$ be the number of  $3$-Sylow subgroups. Now $n_{13}=1+13k$ where $1+13k|9$. Thus the only possible choice for $k$ is $0$. Hence, there is a unique $13$-Sylow subgroup and because it is unique it is normal.
    \end{proof}
\end{minipage}

\caption{A sample theorem statement from mathlib, show on the left, and a sample theorem statement from \textsf{ProofNet}, shown on the right. Mathlib emphasizes including the most abstract and general formulations of mathematical results, whereas \textsf{ProofNet} predominantly tests the ability of models to apply those results to concrete problems.}
\label{fig:style}
\vspace{-10pt}
\end{figure*}

The creation of an automatic mathematician, that is, a system capable of autonomously posing conjectures and proving theorems, is a longstanding challenge in mathematics and artificial intelligence \citep{gelertner1959}. In recent years, neural generative language modelling has emerged as a promising approach to automating aspects of mathematics \citep{rabe2021}. 

One approach to applying language models to mathematics has been to treat mathematical reasoning in natural language as a sequence learning task \citep{welleck2021, welleck2022, lewkowycz2022}. A key advantage of mathematical reasoning in natural language is the abundance of natural language mathematics data on the internet \citep{lewkowycz2022}.

An alternative approach is to use language models to guide formal proof-search in an interactive theorem prover (ITP) \citep{yang, polu2020, polu2022, jiang2022}. A salient advantage of this method is that the ITP acts as a verifier for the language model's reasoning, enabling the natural implementation of bootstrapping techniques such as expert iteration \citep{silver2017, polu2022}. 

{\it Autoformalization}, the task of automatically formalizing mathematics, seeks to build a bridge between informal and formal mathematical reasoning \citep{wang2018, szegedy2020, wu2022}, with the potential of extracting a training signal from vast corpora of natural language mathematics data while still grounding a system's reasoning in formal logic. However, lack of parallel data between informal and formal mathematics means that autoformalization suffers from a lack of standard benchmarks to guide progress in the field. 

To remedy this gap, we propose \textsf{ProofNet},\footnote{Full dataset are available at \url{https://huggingface.co/datasets/hoskinson-center/proofnet}. Code to replicate experiments available at \url{https://github.com/zhangir-azerbayev/ProofNet}} a benchmark consisting of parallel natural language and formal mathematics that can be used to evaluate autoformalization and theorem proving. The \textsf{ProofNet} benchmark consists of 371 parallel formal theorem statements, natural language theorem statements, and natural language proofs sourced from the exercises of popular undergraduate-level pure mathematics textbooks. Formal statements are expressed in the Lean 3 theorem prover \citep{MouraKADR15}, and depend on Lean's {\sf mathlib} \citep{mathlib}. 

Language-model-based theorem provers and autoformalization systems have typically been evaluated on benchmarks consisting of competition and olympiad-style problems \citep{zheng22, wu2022}. While such problems require complex reasoning, their solutions only depend on a relatively small set of elementary facts about integers, real numbers, counting, and geometry. In contrast, modern research mathematics requires the mastery of a massive body of theory made up of thousands of definitions, lemmas, and theorems. The Lean 3 formalization of perfectoid spaces, an important definition in research-level arithmetic geometry, depends on over 3000 distinct theorems and definitions \citep{buzzard2020}. How to effectively reason over such a large repository of knowledge is an important unsolved problem in applying language models to mathematics \citep{irving2016, wu2022mem, tworkowski2022} .

\textsf{ProofNet} falls short of requiring mastery of all of modern mathematics, but poses the still ambitious goal of reasoning over the core of an undergraduate mathematics, including basic analysis, algebra, number theory, and topology. We hope that this benchmark will spur the development of language models that are able to reason effectively over large knowledge bases.

In order to obtain stronger baselines on {\sf ProofNet}, we train and open-source the {\sc proofGPT} language models at scales of 1.3 billion and 6.7 billion parameters. These models are trained on the {\sf proof-pile}, an 8 billion token dataset of mathematical text. To our knowledge, these are the only open-source language models fine-tuned for general mathematics. 

We establish baselines for {\sf ProofNet} theorem autoformalization using in-context learning \citep{brown2020}. Moreover we introduce two novel theorem autoformalization methods that outperform our few-shot baselines. {\it Prompt retrieval} uses nearest-neighbor search against an embedding database to create a prompt consisting of the {\sf mathlib} declarations most relevant to a particular natural language theorem. {\it Distilled backtranslation} is a method inspired by work in unsupervised machine translation \citep{lample2017, han2021} that finetunes a language model for autoformalization at a large scale without the need for parallel data.

\section{The \textsf{ProofNet} Benchmark}

\paragraph{Dataset collection} Problems in the \textsf{ProofNet} benchmark are primarily drawn from exercises in popular undergraduate mathematics textbooks. For a complete list of sources, see \autoref{appendix:sources}.

Not all textbook exercises lend themselves naturally to formalization. In particular, we only consider for inclusion in \textsf{ProofNet} problems meeting the following criteria: 
\begin{itemize}
    \item {\it Self-containment}. Problems should only depend on the results commonly taught in an undergraduate curriculum. In particular, this rules out problems that are split into multiple sequentially dependent parts, or those using nonstandard notations. 
    \item {\it Naturality of formalization.} Not all kinds of mathematical problems can be naturally formalized, such as word problems, and such problems are excluded. We do not include exercises that require computing an unknown quantity. We do not include problems that depend on parts of Lean's mathlib that are relatively less mature, such as Euclidean geometry or combinatorics. 
    \item {\it Low risk of train-test overlap.} Because language models are often pre-trained on large corpora mined from the internet that include {\sf mathlib}, we refrain from including statements that are in mathlib or are likely to be added to mathlib in the future. In practice, this means we avoid the abstract ``theory-building'' style of theorems that constitute mathlib, and instead choose problems that involve applying general results to specific cases. For more insight into the stylistic differences between mathlib and \textsf{ProofNet} problems, see \autoref{fig:style}.
\end{itemize}
Beyond the above criteria, problems were selected for broad coverage of the undergraduate curriculum and to range in difficulty from straightforward applications of the definitions to those requiring tremendous creativity. Problems statements are transcribed into \LaTeX{} and formalized by human annotators proficient in Lean. Natural language proofs are adapted from online solutions manuals, or in a few cases, written by the annotators. 

\paragraph{Supported Tasks} As \textsf{ProofNet} includes parallel natural language statements, 
natural language proofs, and formal statements, the dataset supports the evaluation of the following distinct tasks: 
\begin{itemize}
    \item {\it Formal theorem proving}. Given a formal statement of a theorem, produce a formal proof. 
    \item {\it Informal theorem proving}. Given an informal statement, produce an informal proof.
    \item {\it Autoformalization and informalization of statements}. Given an informal (formal) statement, produce a corresponding formal (informal) statement.
    \item {\it Autoformalization of proofs}. Given an informal theorem statement, its informal proof, and its formal statement, produce a formal proof.
\end{itemize}

\section{The {\sc proofGPT} models}
    \begin{table}[t]
        \centering
        \begin{tabular}{lcc}
            \toprule
            Source & Size (GB) & Tokens \\
            \midrule
            arXiv.math & 13.6 & 4.9B \\
            Stack Exchanges & 0.96 & 0.3B \\
            Formal math libraries & 0.14 & 59M\\
            ProofWiki + Wikipedia math articles & 0.02 & 6.6M \\
            Open source books & 0.015 & 6.5M \\
            MATH & 0.002 & 0.9M\\
            \bottomrule \\ 
        \end{tabular}
        \caption{Composition of the {\sf proof-pile}.} 
        \label{tab:pretrain}
    \end{table}

    \begin{table}[t]
        \centering
        \begin{tabular}{lcc}
            \toprule
            Model & {\sf arXiv.math} perplexity & {\sf proof-pile} perplexity\\
            \midrule
            {\it 1B parameters:} & &\\
            \quad Pythia 1.4B & 3.82& 4.12  \\
            \quad proof-GPT 1.3B & 3.17& 3.47  \\
            \midrule 
            {\it 6B parameters:} & &\\
            \quad Pythia 6.9B & 3.36 & 3.62  \\
            \quad proof-GPT 6.7B &3.12  &3.43\\
            \midrule 
        \end{tabular}
        \caption{Comparison of model perplexities on the test set of the {\sf arXiv} subset of the {\sf proof-pile} and the entire {\sf proof-pile}. Documents were joined using two newline characters and perplexity was calculated with a stride equal to the model's context length, which is 2048 for all models shown.} 
        \label{tab:gpt}
    \end{table}

Many approaches to quantitative reasoning with language models depend on pre-training or fine-tuning a model on large corpora of mathematical text, which significantly boosts downstream performance \citep{hendrycks2021, polu2020, lample2022, lewkowycz2022}. Motivated by these results, we train and open-source the {\sc proofGPT} models at sizes of 1.3 billion and 6.7 billion parameters \footnote{\url{https://huggingface.co/hoskinson-center/proofGPT-v0.1}\\\url{https://huggingface.co/hoskinson-center/proofGPT-v0.1-6.7B}}. The {\sc proofGPT} models are decoder-only causual language models initialized with Pythia weights \citep{pythia}\footnote{The {\sc proofGPT} models were not initialized from the open-sourced weights of the Pythia models, but from a development version of the suite with slightly different hyperparameters. This is the cause of the small parameter discrepancy between a {\sf proofGPT} and the similarly sized Pythia model. Performance of the development versions of Pythia and the open-source versions are near-identical.}, and then fine-tuned on the {\sf proof-pile}, a corpus of mathematical text whose composition is detailed in \autoref{tab:pretrain}. Fine-tuning was performed using the GPT-NeoX library \citep{gpt-neox}. For training hyperparameters, see \autoref{appendix:training}. In \autoref{tab:gpt}, we show that the {\sc proofGPT} models outperform Pythia base models and GPT-2 at standard mathematical reasoning tasks. 

We regard the {\sc proofGPT} model suite as inferior to the Minerva models \citep{lewkowycz2022} due to the fact that the {\sc proofGPT} models are fine-tuned on an order of magnitude less mathematical text and span a smaller parameter range. However, we hope that the research community will benefit from {\sc proofGPT}'s open-source weights and dataset. 

\section{Methodology and Experiments}
In this work, we evaluate the capabilities of pre-trained language models on autoformalizing and informalizing theorem statements. 
Due to the engineering challenges of implementing neural theorem proving systems in Lean, we leave an investigation of formal theorem proving and proof autoformalization to future work. 

\subsection{Autoformalization methods}
We employ in-context learning with large language models as a strong baseline for the autoformalization of theorem statements \citep{wu2022}. Moreover, we introduce two novel methods for boosting autoformalization performance above the few-shot baseline: {\it prompt retrieval} and {\it distilled backtranslation}. 

\subsubsection{Few-shot autoformalization and informalization} In-context learning is a simple and powerful method for adapting language models to sequence-to-sequence tasks \citep{brown2020}.

For our in-context baselines, we perform inference using the OpenAI API's {\it Code-davinci-002} endpoint \citep{chen2021} and the proofGPT 1.3B and 6.7B models. Prompts are listed are given in \autoref{appendix:few-shot}.

Because there may be multiple ways to formalize the same statement in Lean and no general way to automatically verify whether two statements that are not definitionally equal have the same mathematical content, autoformalizations are evaluated for correctness by a human expert. Informalizations are also judged by a human expert. 

\subsubsection{Prompt retrieval} A blessing and a curse of current language models is that few-shot learning performance is highly sensitive to the exact prompt that is used \citep{kojima2022}. In particular, it is plausible that greater few-shot learning performance can be achieved by retrieving the few-shot examples that are most relevant to a particular question. 

We implement a {\it prompt retrieval} procedure for statement autoformalization as follows. Suppose we have a knowledge-base $\mathcal{K}$ of formal statements. First, we generate an autoformalization $\hat{y}$ of a statement $x$ using our standard in-context procedure. Then we produce dense vector representations of $\hat{y}$ and the formal statements in $\mathcal{K}$. We retrieve the $k$-nearest-neighbors of $\hat{y}$ in $\mathcal{K}$, and include them in the few-shot prompt. For the precise format of the prompt, see \autoref{appendix:few-shot}. 

We opt to retrieve against $\hat{y}$ instead of against $x$ because this method was significantly more performant in our preliminary experiments.

In our experiments, we create a knowledge-base $\mathcal{K}$ by taking our $y$s to be 90,530 statements from Lean {\sf mathlib} and use $k=4$. We use the OpenAI API's {\it embedding-ada-002} endpoint to generate text embeddings.

\subsubsection{Distilled backtranslation} A significant obstacle to fine-tuning language models on the autoformalization of theorem statements is the lack of large parallel corpora of formal and informal mathematics. In the face of this limitation, we draw on prior work leveraging generative models for unsupervised translation between natural languages. In particular, we use {\it distilled backtranslation}, a methodology inspired by \cite{han2021}. 

Distilled backtranslation proceeds as follows. Suppose we have a large language model $P_{LLM}(\cdot)$ pre-trained on monolingual data in both the source and target language, a monolingual corpus $\{Y_i\}$ in the target language. We wish to fine-tune a ``student'' model $P_\theta(Y|X)$ to translate a sequence $X$ in the source language to a corresponding sequence $Y$ in the target language. First, we manually construct a few-shot prompt $C$ consisting of $X|Y$ pairs. Then, we sample synthetic backtranslations $X_i \sim P_{LLM}( X | C, Y_i)$. Finally, we fine-tune $P_\theta(\cdot)$ on the synthetic pairs to predict $P(Y|X)$. 

In our experiments, we fine-tune proofGPT-1.3B using distilled backtranslation with informal mathematics as the source language and Lean 3 theorems as the target language. We use the theorems in Lean's mathlib as the target language's monolingual corpus. We use {\it Davinci-codex-002} as our teacher LM and proofGPT-1.3B as our student model. Fine-tuning hyperparameters are described in \autoref{appendix:finetuning}


\section{Results and Discussion}
\begin{table}
  \centering
  
      \begin{tabular}{lcccccc}
      \toprule
              & \multicolumn{3}{c}{\it Formalization} & \multicolumn{3}{c}{\it Informalization}\\
              \cmidrule(r){2-4} \cmidrule(r){5-7}
        Model &  Typecheck rate& BLEU & Accuracy & Compile rate& BLEU & Accuracy\\
        \midrule
        {\it Few-shot.} & & & & & &\\
        proofGPT-1.3B &5.9 & 8.1 &0 &0.77 & 5.1&4.3\\
        proofGPT-6.7B  & 4.3 &4.7  &0  & 0.70  & 6.0 & 6.5 \\
        Codex    & 23.7  & 25.1  &13.4 & 100  & 13.2  & 62.3  \\
        \midrule 
        {\it Prompt retrieval:} & & & & & & \\
        Codex & 45.2 & 14.8 & 16.1 & -&- &- \\
        \midrule 
        {\it Dist. backtrans.} &&&&&\\
        proofGPT-1.3B & 19.4  & 10.7  & 3.2  & - & - & - \\
        \bottomrule\\
      \end{tabular}
    \caption{Results of few-shot learning with LLMs on formalization and informalization of \textsf{ProofNet} statements. In addition to reporting autoformalization accuracy, we also report {\it typecheck rate}, which is the proportion of a model's samples that are well-formed statements in Lean's dependent type theory. If a model simply copies a formal statement from its prompt, we do not consider that a positive sample when calculating typecheck rate. For the informalization task, we also report {\it compile rate}, i.e what proportion of the model's samples produce \LaTeX{} that compiles. The most common reason why informal generations fail to compile is that they contain Unicode characters frequently used in Lean's mathlib but not accepted by the pdflatex compiler. To calculate BLEU scores, we split on whitespace and use BLEU-4 with smoothing. Note that formalization BLEU scores being higher than informalization BLEU scores is likely because natural language contains more lexically distinct but semantically equivalent statements.}
    \label{tab:results}
\end{table}

%

\subsection{In-context learning}
In \autoref{tab:results}, we present our experimental results for autoformalization and informalization of \textsf{ProofNet} theorem statements. Although conceptually simple and easy to implement, our {\it Code-davinci-002} in-context learning baseline achieves highly nontrivial performance, correctly formalizing 13.4\% of theorems. The {\sc proofGPT} models do not formalize any statements correctly, likely owing to their smaller parameter count. However, they demonstrate some signal on the typecheck rate and BLEU metrics. Note that even generating statements that typecheck in Lean 3's strict type theory is a nontrivial feat. 

Informalization accuracy is much higher than formalization accuracy for all models, supporting the intuitive claim that informalization is an easier task than formalization. This result also suggests that large pre-trained language models have a strong grasp of the semantics of formal mathematics, and primarily struggle with generating lexically correct and type correct Lean code. 

We further observe that among {\it Code-davinci-002}'s generations that typecheck, roughly half are correct formalizations. This is consistent with our hypothesis that {\it Code-davinci-002} has a strong grasp of the semantics of mathematics, since the model displays high accuracy conditional on having generated valid Lean.

\subsection{Prompt Retrieval and Distilled Backtranslation}
In \autoref{tab:results}, we additionally include autoformalization scores for the prompt retrieval and distilled backtranslation models. Applying prompt retrieval to the \davinci{} model significantly boosts performance, increasing accuracy by 2.7 points and, notably, increasing typecheck rate by 21.5 points. 

Distilled backtranslation improves the autoformalization performance of the {\sc proofGPT} 1.3B model not merely above the in-context performance of {\sc proofGPT} 1.3b, but also above the in-context learning performance of {\sc proofGPT} 6.7B. 

\paragraph{Automatic metrics} Typecheck rate correlates strongly with formalization accuracy, and we recommend that typecheck rate be used as a predictor of autoformalization performance when evaluating accuracy is too costly. The BLEU metric correlates well with performance on many NLP tasks \citep{bleu}, but correlates poorly with performance code tasks \citep{chen2021}. Our findings illustrate that just as with code, BLEU is a poor guide to formalization performance, as prompt retrieval increases \davinci{} formalization accuracy but decreases BLEU by over 10 points. 

\subsection{Qualitative Analysis}
We ground our remaining analysis in four case studies: two that demonstrate successful formalizations produced by \davinci, and two that are representative of our methods' most common failure cases.

\paragraph{Case study 1.} The following is a correct and idiomatic Lean mathlib formalization of exercise 10.7.4 in Michael Artin's {\it Algebra} produced by \davinci:
\begin{center}
\begin{tabular}{|l|}
\toprule
     NL: Let $R$ be a ring, with $M$ an ideal of $R$. Suppose that every element of $R$ which is not in\\ $M$ is a unit of $R$. Prove that $M$ is a maximal ideal and that moreover it is the only maximal\\ ideal  of $R$.
  \\
\midrule
    {\it Code-davinci-002} output: \\
     \texttt{{\color{blue}theorem} {\color{orange}exercise\_10\_7\_10} \{R : {\color{blue}Type}*\} [ring R] (M : ideal R)} \\
     \texttt{  (hM : $\forall$ x : R, x $\notin$ M $\to$ is\_unit x)} :\\
     \texttt{  is\_maximal M $\wedge$ $\forall$ (N : ideal R), is\_maximal N $\to$ N = M}\\
\bottomrule
\end{tabular}
\end{center}
In this example, \davinci{} demonstrates strong knowledge of {\sf mathlib} and some reasoning capability. For example, the model correctly identifies that ideals are expressed using a typeclass \texttt{ideal R}, but maximal ideals are expressed using a predicate \texttt{is\_maximal}. Moreover, the model correctly recognizes that an instance of the \texttt{ideal} type can be implicitly coerced to a \texttt{set R} in order to apply the set-theoretic notation \texttt{$\notin$}. Finally, the model correctly infers that the best way to express ``only maximal ideal'' is the statement \texttt{$\forall$ (N : ideal R), is\_maximal N $\to$ N = M}, which demonstrates some logical reasoning ability.

\paragraph{Case study 2.} The following problem, exercise 6.4.2 from Michael Artin's {\it Algebra}, is an example of vanilla {\it Davinci-code-002} making a mistake that is rectified by prompt retrieval. 
\begin{center}
\sethlcolor{pink}  
\begin{tabular}{|l|}
\toprule
     NL: Prove that no group of order $pq$, where $p$ and $q$ are prime, is simple.
  \\
\midrule
    {\it Code-davinci-002} output: \\
     \texttt{{\color{blue}theorem} {\color{orange}exercise\_6\_4\_2} \{G : {\color{blue}Type*}\} [group G] [fintype G] \{p q : $\mathbb{N}$\}}\\
     \texttt{(hp : nat.prime p) (hq : nat.prime q) (h : fintype.card G = p * q) :}\\
     \texttt{$\neg$ \hl{simple\_group} G} \\
\midrule 
    {\it Code-davinci-002} + retrieval output: \\
     \texttt{{\color{blue}theorem} {\color{orange}exercise\_6\_4\_2} \{G : {\color{blue}Type*}\} [group G] [fintype G] \{p q : $\mathbb{N}$\}}\\
     \texttt{[hp : fact(nat.prime p)] [hq : fact(nat.prime q)]} \\
     \texttt{(h : fintype.card G = p * q)} :\\
    \texttt{\sethlcolor{green}$\neg$ \hl{is\_simple\_group} G} \\
\bottomrule
\end{tabular}
\end{center}
Vanilla {\it Davinci-code-002} makes a single lexical error: the model has correctly formalized the hypotheses of the problem, but has tried to use a nonexistent predicate \texttt{simple\_group} instead of the correct \texttt{is\_simple\_group}. The retrieval-augmented model uses the correct declaration because one of the retrieved {\sf mathlib} statements explicitly uses \texttt{is\_simple\_group}. 

\paragraph{Case study 3.} The following is an incorrect formalization of exercise 6.16 from Sheldon Axler's {\it Linear Algebra} produced by \davinci. The retrieval-augmented \davinci{} autoformalization, not shown here, is also incorrect. 
\begin{center}
\sethlcolor{pink}  
\begin{tabular}{|l|}
\toprule
     NL: Suppose $U$ is a subspace of $V$. Prove that $U^{\perp}=\{0\}$ if and only if $U=V$
  \\
\midrule
    {\it Code-davinci-002} output: \\
     \texttt{{\color{blue}theorem} {\color{orange}exercise\_6\_16} \{K : {\color{blue}Type}*\} [field K]} \\
     \texttt{  \{V : {\color{blue}Type}*\} [add\_comm\_group V] [\hl{vector\_space K V}] \{U : \hl{subspace V}\} :}\\
     \texttt{  U.\hl{orthogonal} = \hl{\{0\}} $\leftrightarrow$ U = $\top$}\\
\midrule 
    {\it ProofNet} label: \\
    \texttt{{\color{blue}theorem} {\color{orange}exercise\_6\_16} \{K : {\color{blue}Type}*\} [is\_R\_or\_C K] [inner\_product\_space K V] }\\
     \texttt{  (U : submodule K V) :}\\
     \texttt{  U.orthogonal = $\bot \leftrightarrow$ U = $\top$}\\

\bottomrule
\end{tabular}
\end{center}
Here, \davinci{} makes three straightforward lexical mistakes: in {\sf mathlib}, vector spaces over a field $K$ are denoted by \texttt{module K V} rather than \texttt{vector\_space K V}, subspaces of a vector space by \texttt{submodule K V} rather than \texttt{subspace V}, and the trivial subspace is denoted by \texttt{$\bot$} rather than \texttt{\{0\}}. However, the model also makes a much more significant logical error. In order for the orthogonal complement $U^\perp$ of a subspace to make sense, the space must be endowed with a {\it inner product space} structure rather than merely a vector space structure, which is expressed in {\sf mathlib} as \texttt{inner\_product\_space K V}. Furthermore, inner product spaces are only defined over the real and complex fields, so one must also declare \texttt{[is\_R\_or\_C K]}. Reliably inferring these kinds of implicit hypotheses is a major challenge for autoformalization systems.

\paragraph{Case Study 4: Looping.} When generating formal and natural language theorem statements with {\sc proofGPT}s, the model will often repeatedly generate the same short phrase instead of producing a plausible theorem. For example, consider the attempted formalization of exercise 10.1.13 from Michael Artin's {\it Algebra} generated by {\sc proofGPT} 6.7B via in-context learning.
\begin{center}
\begin{tabular}{|l|}
\toprule
     NL: An element $x$ of a ring $R$ is called nilpotent if some power of $x$ is zero.\\ Prove that if $x$ is nilpotent, then $1+x$ is a unit in $R$.
  \\
\midrule
    {\it proofGPT-6.7b} output: \\
\texttt{{\color{blue}theorem} nilpotent\_of\_nilpotent\_of\_nilpotent\_of\_nilpotent\_of\_nilpotent}\\
\texttt{nilpotent\_of\_nilpotent\_of\_nilpotent\_of\_nilpotent\_of\_nilpotent\_of$\dots$}\\ 
\bottomrule
\end{tabular}
\end{center}
Prior work on decoding methods has shown that the likelihood of a repeated phrase increases with each repetition, and that greedy decoding generates text with higher likelihood than natural text \citep{holtzman}. These two findings constitute a plausible explanation for repetitive looping if the correct autoformalization is assigned low likelihood by the model. We observe that repetitive looping does not occur with {\it Code-davinci-002}, suggesting that the problem may disappear with scale (although there are many other differences between our small-scale models and {\it Code-davinci-002}).

\section{Related Work}

\paragraph{Language modeling for theorem proving} Language models have found success in theorem proving both in the natural language setting \citep{lewkowycz2022, welleck2021}, and within many major ITPs such as Metamath \citep{polu2020}, Isabelle \citep{jiang2022}, and Lean \citep{han2021pact, polu2022}. Popular benchmarks for evaluating language model-based provers are \cite{hendrycks2021} and \cite{welleck2021} for natural language, and \cite{zheng22} for formal. 

\paragraph{Autoformalization} Recent work in autoformalization with language models was sparked by \citet{wu2022}, which demonstrated that models can autoformalize Isabelle theorem statements via in-context learning. In \citet{dsp}, the authors demonstrate a method for autoformalizing proofs in Isabelle. However, their method depends on the availibility of a performant black-box automated theorem prover, which is not available for Lean at the time of writing. 

\paragraph{Interactive Theorem Proving} Work in formal theorem proving and autoformalization depends on libraries of formalized mathematics. This work directly depends on Lean's {\sf mathlib}, but indirectly benefits from lessons learned from other proofs systems such as Coq \citep{BertotC04}, Mizar \citep{GrabowskiKN10}, and Isabelle \citep{NipkowPW02}. 

\paragraph{Unsupervised Machine Translation} Because the amount of parallel formal and natural language text is negligible, autoformalization faces many of the same challenges as unsupervised machine translation \citep{lample2017, han2021, garcia}. Our distilled backtranslation method is inspired by the distilled and iterated backtranslation algorithm of \citet{han2021}. However, the authors of this work regard backtranslation as a temporary hack and foresee that in-context learning will be enough to elicit maximal performance from a sufficiently good language model, as is now the case for unsupervised translation \citep{garcia}. 
\section{Conclusion}
We introduced \textsf{ProofNet}, a benchmarking consisting of parallel natural language theorem statements, natural language proofs, and formal theorem statements in Lean 3. We have shown that pre-trained large language models achieve non-trivial but far from consistent performance via in-context learning on the autoformalization of \textsf{ProofNet} statements. Moreover, we have proposed prompt retrieval and distilled backtranslation, two methods that improve autoformalization performance above baseline. 

\section*{Acknowledgments}

The authors would like to thank the Hoskinson Center for Formal Mathematics at Carnegie Mellon University for its generous funding and for providing a stimulating work environment. We additionally thank EleutherAI for providing compute to train the {\sc proofGPT} models. 
Piotrowski was supported also by the grant of National Science Center, Poland, no. 2018/29/N/ST6/02903, and by the Kosciuszko Fundation.

\bibliographystyle{plainnat}
\bibliography{refs}


\appendix

\section{{\sc proofGPT} training}
\label{appendix:training}
\autoref{tab:gptparams} displays hyperparameters for {\sc proofGPT} training on the {\sc proof-pile}. 
\section{Problem Sources}
\label{appendix:sources}
The following is a complete list of sources \textsf{ProofNet} draws from:
\begin{itemize}
    \item Analysis: Walter Rudin's {\it Principles of Mathematical Analysis} 3rd ed, Charles C. Pugh's {\it Real Mathematical Analysis} 1st ed, Elias M. Stein and Rami Shakarchi's {\it Complex Analysis} 1st ed. 
    \item Linear Algebra: Sheldon Axler's {\it Linear Algebra Done Right} 2nd ed. 
    \item Abstract Algebra: David S. Dummit and Richard M. Foote's {\it Abstract Algebra} 3rd ed, I.N. Herstein's {\it Abstract Algebra} 3rd ed, and Michael Artin's {\it Algebra} 1st ed.
    \item Topology: James Munkres' {\it Topology} 2nd ed. 
    \item Examinations: Putnam Competition. 
\end{itemize}

\section{Prompts}
\label{appendix:few-shot}

Prompts are viewable in the open-source repository\footnote{\url{https://github.com/zhangir-azerbayev/ProofNet/tree/main/eval/prompts}}. We use a 12-shot prompt for \davinci{} autoformalization and informalization, and a 6-shot prompt for {\sf proofGPT} autoformalization and informalization. We give {\sf proofGPT} models fewer examples because of its shorter context (2048 tokens compared to 8192), we only use the last six examples when prompting {\sf proofGPT}. 

For retrieval augmented models, we use a 3-shot prompt, where each example consists of 4 reference formal statements and one NL-formal pair. 
\section{Finetuning}
\label{appendix:finetuning}
Our fine-tuning dataset of backtranslations consists of 90,530 NL-formal pairs. Both the Pythia-1.4b and \textsc{proofGPT}-1.3B model are finetuned according to the hyperparameters above. The models evaluated in \autoref{tab:results} are the minimum validation loss checkpoint, which occurs at 15,000 training steps. 

\begin{table*}[t]
\centering
\begin{tabular}{lcc}
\toprule
& \multicolumn{2}{c}{Setting}\\
\cmidrule(r){2-3}
Parameter & 1.3B & 6.7B\\
\hline 
Tokens & \multicolumn{2}{c}{10.5 billion}\\
Epochs & \multicolumn{2}{c}{1.3}\\
Training Steps & \multicolumn{2}{c}{40,000}\\
Learning Rate Max & $2\cdot 10^{-4}$ & $1.2\cdot 10^{-4}$ \\
Learning Rate Min & $2\cdot 10^{-5}$ &  $1.2\cdot 10^{-5}$ \\
Optimizer & \multicolumn{2}{c}{Adam}\\
Adam Betas & \multicolumn{2}{c}{$(0.9, 0.95)$}\\
Adam Eps & \multicolumn{2}{c}{$1\cdot 10^{-8}$}\\
Weight Decay & \multicolumn{2}{c}{$0.1$}\\
LR Scheduler & \multicolumn{2}{c}{Cosine w/ warm-up}\\
LR Warm-up Steps & \multicolumn{2}{c}{400}\\
Effective Batch Size & \multicolumn{2}{c}{128}\\
Precision & \multicolumn{2}{c}{FP16}\\
Gradient Clipping & \multicolumn{2}{c}{1.0}\\
\bottomrule
\end{tabular}
\caption{{\sc proofGPT} training hyperparameters.}
\label{tab:gptparams}
\end{table*}

\begin{table*}[t]
\centering
\begin{tabular}{c|c}
Parameter & Setting\\
\hline 
Training Steps & 20,000\\
Learning Rate (LR) & $5\cdot 10^{-5}$\\
Optimizer & AdamW\\
Adam Betas & $(0.9, 0.999)$\\
Adam Eps & $1\cdot 10^{-8}$\\
Weight Decay & $0.1$\\
LR Scheduler & Cosine w/ warm-up\\
LR Warm-up Steps & 2000\\
Effective Batch Size & 24\\
Precision & FP16\\
Gradient Clipping & 1.0\\
\end{tabular}
\caption{Student training hyperparameters.}
\label{tabparams}
\end{table*}

\end{document}